\documentclass[letterpaper, 10 pt, conference]{ieeeconf}  

\IEEEoverridecommandlockouts                              

\title{\LARGE \bf
Spatio-temporal Attention Model for Tactile Texture Recognition
}

\author{Guanqun Cao, Yi Zhou, Danushka Bollegala and Shan Luo
\thanks{All the authors are at the Department of Computer Science, University of Liverpool, Liverpool L69 3BX, U.K. Emails:\{g.cao, y.zhou71, danushka, shan.luo\}@liverpool.ac.uk.}}%

\usepackage{ifpdf}
\usepackage{graphicx}
\ifpdf

\graphicspath{{Figs/}{Figs/PDF/}}
\else
\graphicspath{{Figs/}}
\fi
\usepackage{hhline}
\usepackage{color}
\usepackage{calc}
\let\labelindent\relax
\usepackage{enumitem}
\usepackage{bm,amsmath}

\usepackage{nomencl}
\usepackage{multirow}
\usepackage{url}
\usepackage{nomencl}
\usepackage{multirow}
\usepackage{hyperref}
\hypersetup{
	colorlinks = true,
	linkcolor = blue,
	citecolor = blue
}
\usepackage{subcaption}
\usepackage{array}
\usepackage{booktabs}
\newcolumntype{P}[1]{>{\centering\arraybackslash}p{#1}}

\usepackage[flushleft]{threeparttable}
\usepackage{cite}

\usepackage{times}
\usepackage{epsfig}
\usepackage{graphicx}
\usepackage{amsmath}
\usepackage{amssymb}

\begin{document}

\maketitle
\thispagestyle{empty}
\pagestyle{empty}

\begin{abstract}

Recently, tactile sensing has attracted great interest in robotics, especially for facilitating exploration of unstructured environments and effective manipulation. A detailed understanding of the surface textures via tactile sensing is essential for many of these tasks.
Previous works on texture recognition using camera based tactile sensors have been limited to treating all regions in one tactile image or all samples in one tactile sequence equally, which includes much irrelevant or redundant information.
In this paper, we propose a novel Spatio-Temporal Attention Model (STAM) for tactile texture recognition, which is the very first of its kind to our best knowledge. 
The proposed STAM pays attention to both spatial focus of each single tactile texture and the temporal correlation of a tactile sequence.
In the experiments to discriminate 100 different fabric textures, the spatially and temporally selective attention has resulted in a significant improvement of the recognition accuracy, by up to 18.8\%, compared to the non-attention based models. Specifically, after introducing noisy data that is collected before the contact happens, our proposed STAM can learn the salient features efficiently and the accuracy can increase by 15.23\% on average compared with the CNN based baseline approach.
The improved tactile texture perception can be applied to facilitate robot tasks like grasping and manipulation.

\end{abstract}

\section{Introduction}
The sense of touch is one of the important information sources for both humans and robots to perceive the object properties in the physical world. One of the key object properties is the surface texture and the determination of the surface textures is important for object recognition and dexterous manipulation of objects. One of the good examples of the surface textures is the patterns of fabric or clothing. Humans are able to recognise the fabric textures with ease as a result of interaction between the fabric and human skin~\cite{das2011improving}. 
To have robots assist our daily life such as sorting clothes for laundry, it is also important to understand the properties of clothing for service robots.
If robots are able to distinguish whether a fabric is made by cotton or silk through distinguishing their surface textures, clothes can be better sorted, washed and maintained.

\begin{figure}
	\centering
	\includegraphics[width=1\columnwidth]{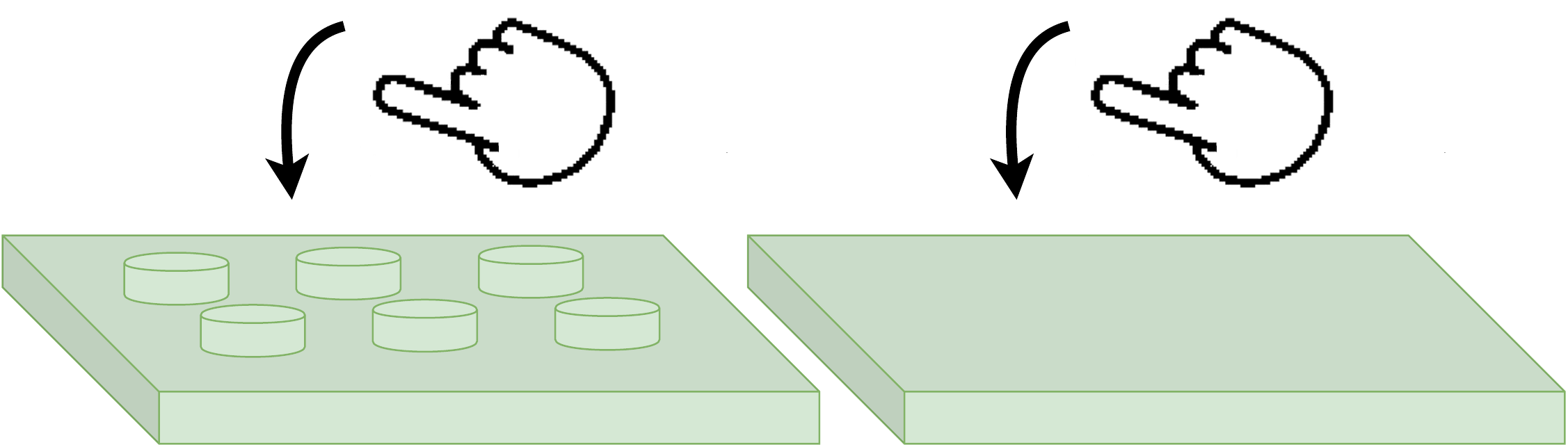}
	\caption{\textbf{\textit{Tactile selective attention.}} If pressing against a Lego brick to distinguish what it is, the studs on the front side (left) that have more distinctive features (e.g., the number/type of the studs) provide more cues than the flat region on the back side that is common in most of the bricks (right).}
	\label{fig:lego}
\end{figure}

Tactile sensors have been used to discriminate surface textures to enable robots have the sense of touch \cite{luo2017robotic,luo2018vitac}. Similar to video sequences collected by a vision camera, a tactile sensor also collects information over a period of time. 
A series of tactile data can be collected by a tactile sensor while the sensor interacts with object surfaces.
When a tactile sensor scans a texture, physical features in the stimulus are concatenated in a temporal order.
The temporal and spatial patterns embedded in tactile sequences are crucial to interpreting the stimulus of surface textures~\cite{gao2016deep}.  

Humans perceive the surface textures by both temporal and spatial patterns presented in tactile sequences~\cite{lederman2009haptic}. 
When we use our fingers to scan an object's surface, i.e., exploring the surface with lateral motions, both spatial and temporal changes in skin deformation provide important cues for fine texture perception~\cite{bensmaia2005pacinian}. 
In this exploratory procedure, we experience the \textbf{\textit{tactile selective attention}}~\cite{sathian1991role}:
in the perceptual area of fingers, we pay our attention to the points that give more excitement rather than treating the whole contacting region equally.
For instance, to distinguish a Lego brick as illustrated in Fig.~\ref{fig:lego}, 
the front side with studs provides more cues (e.g., the number and the type of the studs) than the back side that is common in most of the bricks.
On the other side, perception is an accumulation of cognition that the previous contact events enable a prior knowledge for the perception and later contacts verify the previous judgement. It also means that the contact events are paid attention to different extents while perceiving an object.

Compared to the popularity of the attention-based models in other fields, e.g., the sequence-to-sequence translation in Natural Language Processing (NLP)~\cite{bahdanau2014neural} and image caption generation with visual attention~\cite{xu2015show}, the attention mechanism similar to the human tactile selective attention has not yet been explored in robotics. 
In this paper, we propose a novel Spatio-Temporal Attention Model (STAM) for tactile texture recognition and we believe this is the first work that investigates the attention mechanism in robotic tactile perception. 
We implement the attention model in a task of fabric texture recognition, with 100 pieces of fabrics used. 
The experiments show that our STAM boosts the texture recognition accuracy to a large extent, especially when the data is noisy, compared to the non-attention based models.

The contributions of this paper are as follows:

\begin{enumerate}
    \item We investigate the attention mechanism in the robotic tactile perception, for the first time;
    \item We develop a spatio-temporal attention model that attends to salient features in both spatial and time dimensions of tactile perception;
    \item{A set of experiments demonstrate our proposed method improves tactile texture recognition significantly, which is promising to facilitate manipulation tasks in hand. }
\end{enumerate}


\section{Related Works}\label{sec:relatedwork}
In this section, we will first review works on tactile texture perception, followed by a discussion of the applications of attention models in different domains.

\subsection{Tactile Texture Perception}\label{methods}
Textures have played a key role in understanding properties as they convey important surface characteristics and appearance of an object, given by the shape, size, density, proportion and arrangement of its elementary parts. Different from observing the object textures from a distance in visual texture perception, tactile sensors have direct interaction with the object surface textures~\cite{luo2017robotic}. Various approaches have been proposed in the literature to retrieve the texture information from the collected tactile data. In accordance with the tactile sensors used, these approaches can be categorized into tactile image-based, sequential tactile data-based and spatio-temporal-based approaches.

\textbf{\textit{Tactile image-based texture perception.}} Tactile array sensors such as Weiss tactile sensors~\cite{luo2015novel,madry2014st,luo2019iclap,luo2015localizing,luo2016iterative,luo2015tactile} and optical tactile sensors GelSight~\cite{yuan2017gelsight} and TacTip~\cite{ward2018tactip}, can sense the micro-structure patterns of object textures from the collected tactile images (similar to the visual images). In~\cite{li2013sensing}, height maps of the pressed surfaces collected from a GelSight sensor are used to discriminate surface textures using adapted Local Binary Pattern (LBP) descriptors. In~\cite{luo2018vitac,lee2019touching}, deep learning models were applied to extract texture features from GelSight tactile images and visual images. Similarly, another camera-based tactile sensor TacTip has also been used to analyze the object textures~\cite{ward2018tactip}. 
In such approaches, only the tactile patterns are used to discriminate textures, regardless of the temporal information.

\textbf{\textit{Sequential data-based tactile texture perception.}} 
Most prior approaches use strain gauges or force sensors to detect vibrations during object-sensor interaction, to discriminate surface textures. A BioTac sensor is used in~\cite{fishel2012bayesian} for identification of textures with Bayesian exploration that selects optimal movements based on previous tactile sequence data.
In~\cite{gao2016deep}, the haptic signals from a BioTac sensor are fed into a Convolutional Neural Network (CNN) that performs temporal convolutions, combined with a visual CNN model for multi-modal learning. The sequential tactile data such as induced vibration intensities can also be transferred into the frequency domain~\cite{jamali2010material}.
The surface characteristics like frictions can be revealed from such tactile temporal analysis. However, due to the use of single contact tactile sensors such as strain gauges or force sensors, the local contact patterns cannot be included in discriminating surface textures.

\textbf{\textit{Spatio-temporal tactile perception.}} There have also been prior works on using spatio-temporal tactile features for robot perception tasks. Soh et al.~\cite{soh2012online} proposed an on-line generative model using a sparse Gaussian Process to learn spatio-temporal features from tactile data collected by an iCub robot. In~\cite{madry2014st}, unsupervised hierarchical feature learning is applied to extract features from sequences of raw tactile readings. The learned features are then used for facilitating grasping and object recognition tasks. 
In~\cite{Yuan2018ActiveLearning}, tactile sequences are fed into neural networks to identify the materials of clothing through tactile properties such as thickness, softness and durability. 

\subsection{Applications of Attention Mechanism  }\label{temporal}
Attention mechanism was first proposed to improve the performance of machine translation~\cite{bahdanau2014neural}. Since then, it has been popular in solving various problems in the field of NLP. Attention based models have achieved the state-of-the-art performance in different tasks like abstractive sentence summarization~\cite{rush-chopra-weston:2015:EMNLP}. Contextualized text representation methods, which take the advantage of attention mechanism, such as the Bidirectional Encoder Representations from Transformers (BERT)~\cite{BERT}, yield promising performance for many NLP tasks.
The attention mechanism is usually applied to Recurrent Neural Networks (RNN) and Long Short Term Memory (LSTM) to emphasize salient hidden states for sequential prediction tasks~\cite{larochelle2010learning}. 
Recently, several visual attention models~\cite{xiao2015application,xu2015show} have been proposed. These models are able to automatically locate the discriminated regions in order to better capture differences between images~\cite{zhao2017diversified}. 
However, to the best of the authors' knowledge, there have not been prior works investigating the selective attention in tactile texture recognition.


\begin{figure*}[t]
	\centering
	\includegraphics[width=1.8\columnwidth]{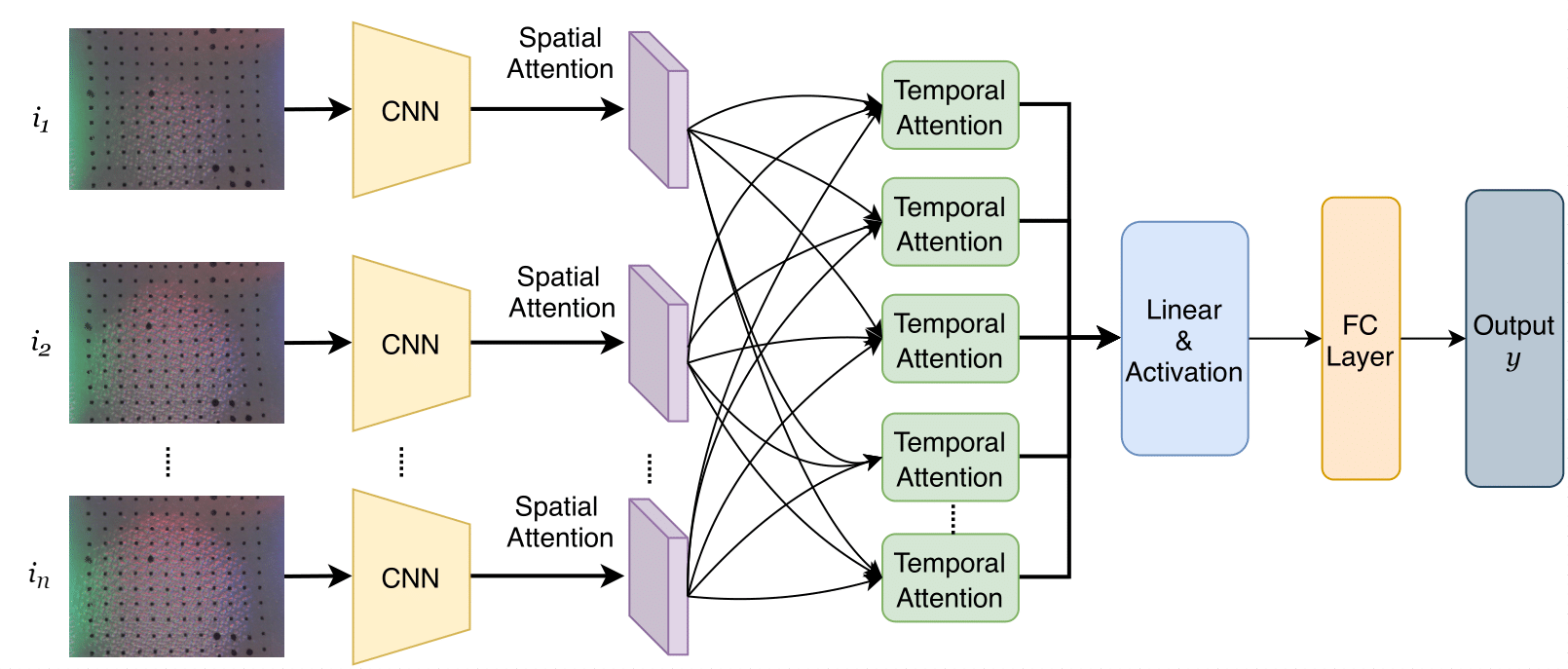}
	\caption{\textbf{\textit{The proposed STAM framework for tactile texture recognition.}} The model receives a sequence of tactile images as input. Each tactile image is fed into a CNN to extract spatial features that is followed by a spatial attention layer to assign weights to different regions. Outputs of the spatial attention layers are then fed into temporal attention modules to learn temporal features. 
   Finally, after the fully connection layers, the STAM model outputs a predicted texture label $y$.} 
	\label{fig:framework}
\end{figure*}

\section{Methodologies}\label{sec:methodology}
In this section, our proposed Spatio-Temporal Attention Model (STAM) is introduced. The STAM receives a sequence of tactile images $I=\{i_1, i_2, ..., i_n\}$ as input, where $ n $ is the length of the tactile sequence. The sequential tactile images are collected from interactions, e.g., pressing, twisting and slipping, between a camera based tactile sensor and an observed object. The STAM outputs a predicted label $y$ which refers to the category of the contacted surface texture. As illustrated in Fig.~\ref{fig:framework}, the STAM model consists of three parts:
1) CNNs that extract spatial features from each input tactile image; 
2) A spatial attention module which highlights the salient features and simultaneously suppresses trivial features in each tactile texture; 3) Temporal attention modules which are used to model the correlation of salient features in different tactile images in one sequence.

\subsection{CNN Module}
Following~\cite{luo2018vitac}, each of the tactile images in the tactile sequence $I$ is first fed into a pre-trained AlexNet architecture~\cite{krizhevsky2012imagenet} simultaneously to extract the spatial features. AlexNet consists of five convolutional layers, the first, second and fifth of which are followed by a max-pooling layer respectively, and three fully connected layers are added on top of the network to output the predicted label. We take the output feature map $\boldsymbol{F} \in \mathbb{R}^{h \times w \times c}$ from the last max-pooling layer of the AlexNet as the input to the spatial attention module, where $h, w, c$ refer to the height, width and the number of channels of the output feature map respectively.

\subsection{Spatial Attention Module}

\begin{figure}
	\centering
	\includegraphics[width=1.0\columnwidth]{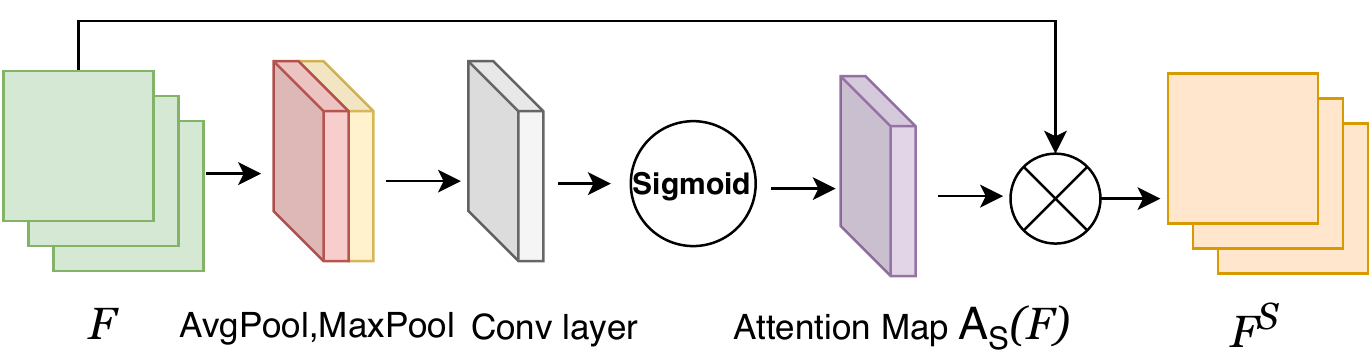}
	\caption{\textbf{\textit{Spatial Attention Module.}} We apply an Average Pooling (AvgPool) and a Max Pooling (MaxPool) 
	on $\boldsymbol{F}$ to get $\boldsymbol{F}_{avg}^S$and $\boldsymbol{F}_{max}^S$. They are followed by a convolutional layer and sigmoid activation to generate an attention map, describing which region containing informative information in $\boldsymbol{F}$. $\otimes$ in the figure represents element-wise multiplication. } 
	\label{fig:spatial}
\end{figure}

In order to emphasize informative areas in each texture frame, we develop a spatial attention module to assign higher weights to crucial areas, whereas lower weights are assigned to the areas that contain less information. The architecture of spatial attention is illustrated in Fig.~\ref{fig:spatial}. Inspired by~\cite{woo2018cbam}, we apply two pooling operations, i.e., max-pooling and average-pooling, to the spatial feature $\boldsymbol{F}$ obtained from the CNN module along the channel axis to form spatial context descriptors. The average-pooling is applied to learn tactile information effectively (with output $\boldsymbol{F}_{max}^S$) while max-pooling is adopted to maintain prominent features (with output $\boldsymbol{F}_{avg}^S$). 
$\boldsymbol{F}_{max}^S$ and $\boldsymbol{F}_{avg}^S$ are then concatenated and convolved with a $7\times7$ kernel, and activated by a sigmoid function to produce a 2D spatial attention map $\boldsymbol{A_{S}(F)}$:
\begin{equation} \label{pooling}
  \begin{split}
 A_{S}(F) &=\sigma (f^{7\times7}([MaxPool(\boldsymbol{F}); AvgPool(\boldsymbol{F})])) \\
 &=\sigma (f^{7\times7}([\boldsymbol{F}_{max}^S ; \boldsymbol{F}_{avg}^S])),
 \end{split}
\end{equation}
where $ \sigma $ denotes the sigmoid function. 
Then we get the output feature map $\boldsymbol{F^{S}} = \boldsymbol{A_{S}(F)}\otimes\boldsymbol{F}$ from the spatial attention module, where $\otimes$ refers to the element-wise multiplication.



\subsection{Temporal Attention Module}
\begin{figure}
	\centering
	\includegraphics[width=1.0\columnwidth]{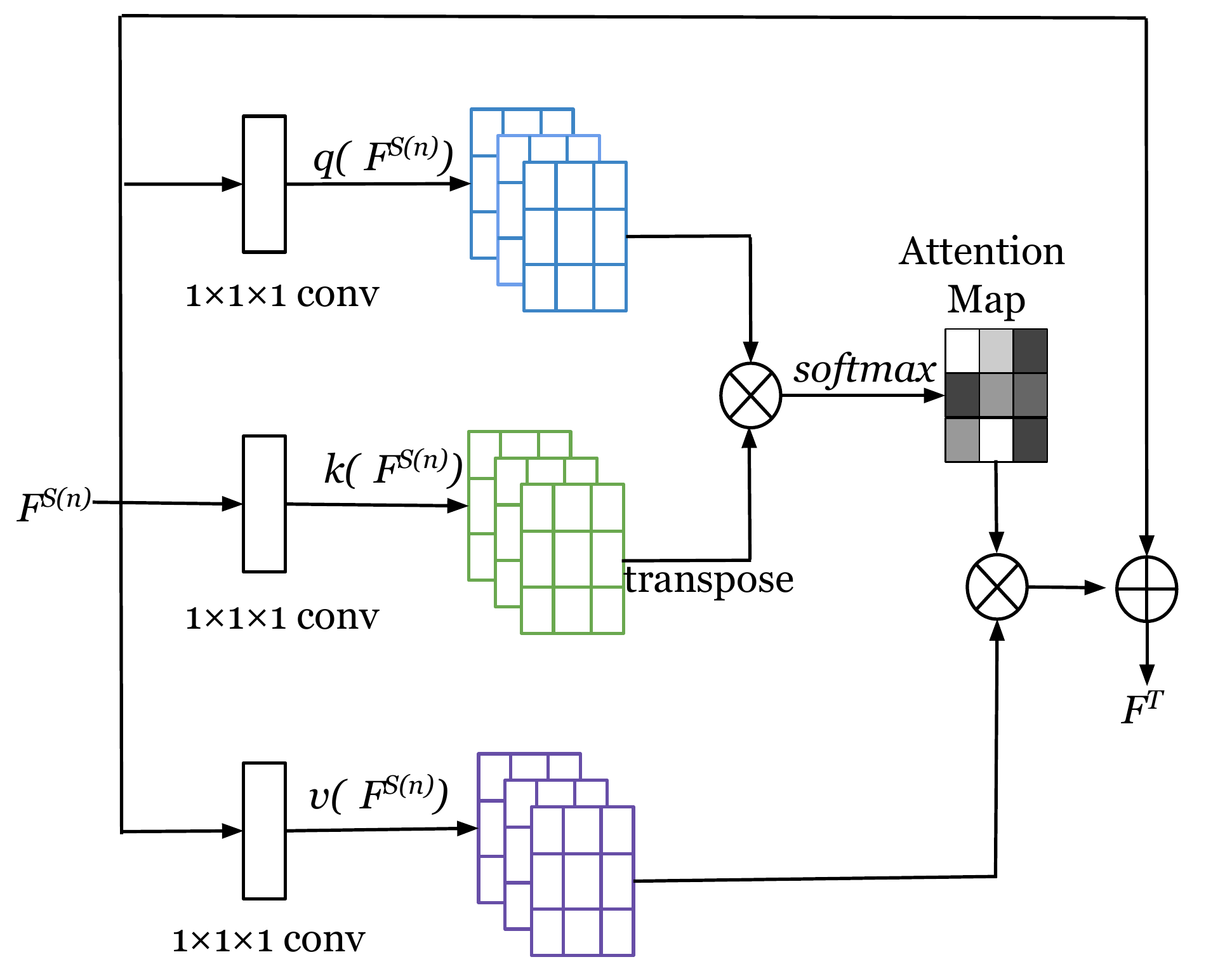}
	\caption{\textbf{\textit{Temporal Attention Module.}} 
	$1\times1\times1$ convolutions transform $\boldsymbol{F^{S(n)}}$ into different feature spaces. 
	The attention map describes the correlation of any pairs of regions in a tactile sequence. 
     $\oplus$ represents element-wise addition.
	}
	\label{fig:temporal}
\end{figure}

After obtaining the extracted features from spatial attention module of each texture frame,
we concatenate all the features together to achieve a sequence of spatial features which can be represented by $\boldsymbol{F^{S(n)}}$. 
In order to model the long-distance dependency in tactile sequence, we develop a temporal attention module 
on top of the spatial attention layer. As illustrated in the Fig.~\ref{fig:temporal}, the module aims to estimate the salience and relevance of all the regions in a tactile sequence through the time regardless of distance. 
$\boldsymbol{F^{S(n)}}$ is first converted into two feature spaces $q(\boldsymbol{F^{S(n)}})$ and $k(\boldsymbol{F^{S(n)}})$ by two sets of $1\times1\times1$ convolutions, where $q(\boldsymbol{F^{S(n)}}) = W_q\boldsymbol{F^{S(n)}}$ and   $k(\boldsymbol{F^{S(n)}}) = W_k\boldsymbol{F^{S(n)}}$ ($ W_q$ and $ W_k$ are trainable weight matrices). 
Subsequently, we reshape both $q(\boldsymbol{F^{S(n)}})\;\text{and}\; k(\boldsymbol{F^{S(n)}}) \in \mathbb{R}^{m \times c}$, where $m = {n}\times{h}\times{w}$, to calculate the attention map of any pairs of regions through time dimension.
The attention map $\boldsymbol{A_T(F^{S(n)})}$  is given as follows: 

\begin{equation} \label{self2}
\boldsymbol{A_T(F^{S(n)})}_{j, i}=\frac{\exp \left(s_{ij}\right)}{\sum_{i=1}^{m} \exp \left(s_{ij}\right)},
\end{equation}
where $s_{ij} = q\left(F^{S(n)}_{i}\right) k\left(F^{S(n)}_{j}\right)^{T}$. $\boldsymbol{A_T(F^{S(n)})}_{j, i}$ demonstrates how much $F^{S(n)}_{i}$ correlates with $F^{S(n)}_{j}$.
The output feature map of the temporal attention is $\boldsymbol{F^{T}} = {\boldsymbol(F^{T}_{1}, F^{T}_{2},...,F^{T}_{j},..., F^{T}_{m})}$,  where
\begin{equation} \label{selfself}
\boldsymbol{\boldsymbol{F^{T}_{j}}}=\sum_{i=1}^{m} \boldsymbol{A_T(F^{S(n)})}_{j, i} \boldsymbol{v}\left(\boldsymbol{\boldsymbol{F^{S(n)}_{i}}}\right)+\boldsymbol{F^{S(n)}_{j}}
\end{equation}
 $\boldsymbol{v}\left(\boldsymbol{F^{S(n)}}\right) = W_v\boldsymbol{F^{S(n)}}$ ($ W_v$ is a learnable matrix) and $\boldsymbol{F^{S(n)}_{j}}$ is added back to keep more information. 

We include multiple temporal attention modules, as shown in Fig.~\ref{fig:framework}, to allow the model to synthesize the information jointly from different representation feature spaces~\cite{vaswani2017attention}.
In the end, the learned hidden representations at the last layer are fed into a fully connected layer to perform a classification task that calculates the probability of a predicted label $y$.

\section{GelSight Sensor and Dataset}\label{sec:datacollection}
In this section, both the GelSight sensor we used~\cite{yuan2017gelsight} and the collected dataset are introduced.
\subsection{The GelSight Sensor}
\begin{figure}
	\centering
	\includegraphics[width=1.0\columnwidth]{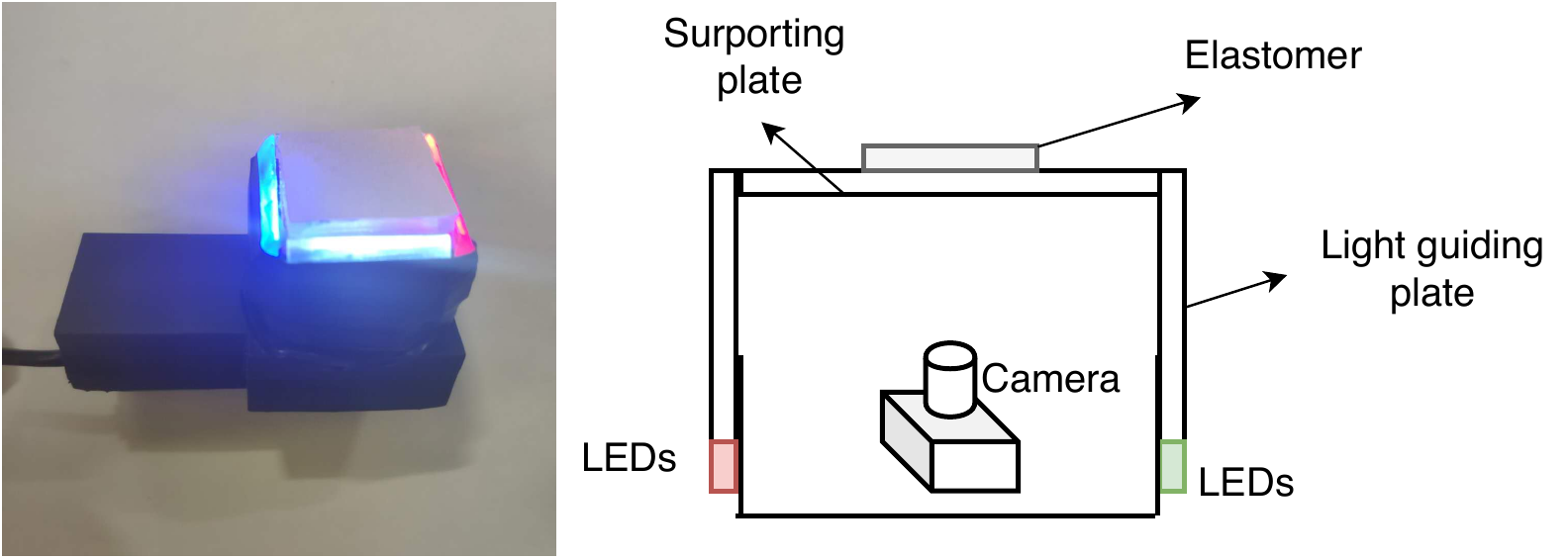}
	\caption{\textbf{\textit{GelSight Sensor.}} Left: a GelSight Sensor. Right: the GelSight structure that consists of a camera at the bottom, a elastomer on a supporting plate to map the object geometry, LEDs and light guiding plates to illustrate the space between the elastomer and the camera. }
	\label{fig:sensor}
\end{figure}

As illustrated in Fig.~\ref{fig:sensor}, the GelSight sensor is a camera based tactile sensor that uses a webcam of a resolution $960 \times 720$ at the base to capture the deformations of the gel layer on the top. The gel layer is made of a piece of clear elastomer coated with reflective membrane. When the GelSight sensor interacts with an object, the geometry of the object surfaces can be mapped to the deformations of the membrane. The camera captures the memberane deformation under illumination from embedded LEDs. To improve the performance of perception in the spatial acuity, multiple square markers with side length of around 0.4 mm are evenly distributed on the elastomer membrane. Readers are referred to~\cite{yuan2017gelsight} for more details of the GelSight sensor.

\subsection{The ViTac Dataset}
\begin{figure*}
	\centering
	\includegraphics[width=1.9\columnwidth]{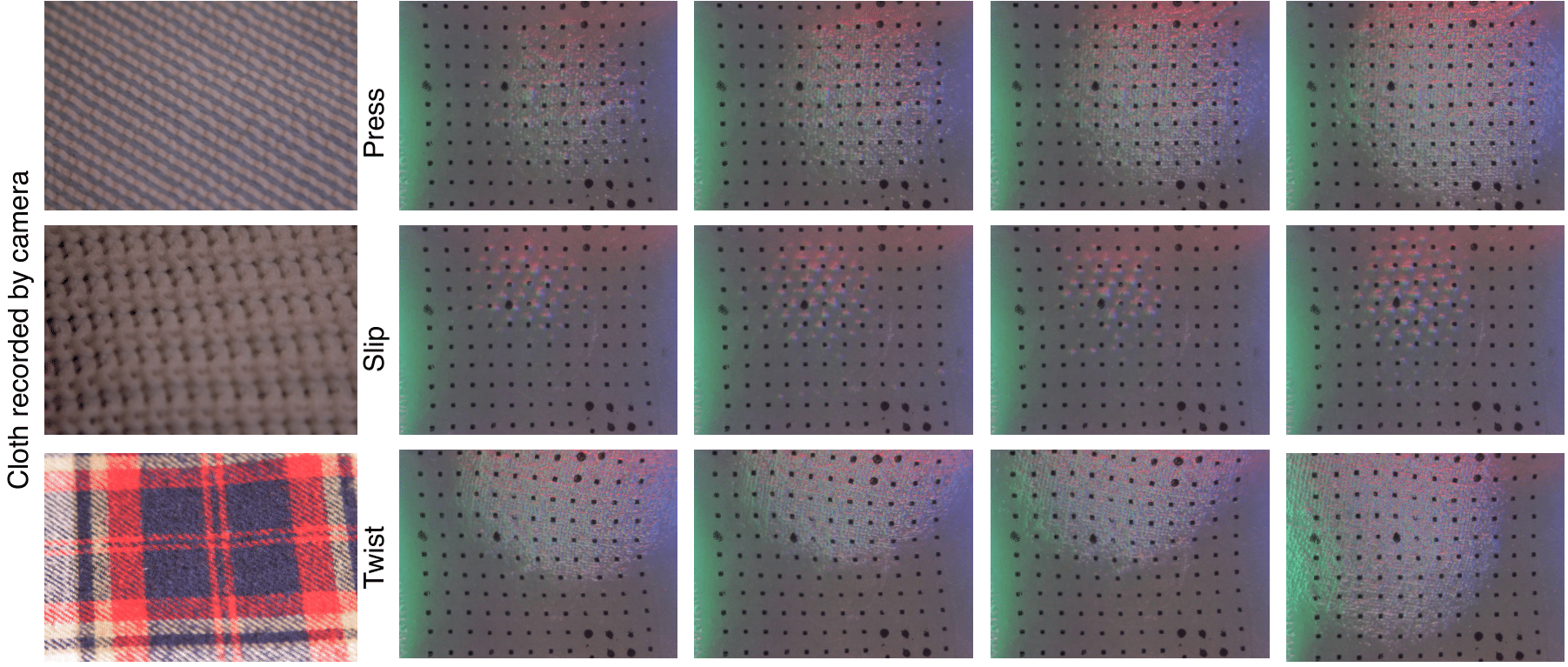}
	\caption{\textbf{\textit{Sample Tactile Sequences.}} Leftmost column: visual images of the fabric samples captured by a digital camera; Right four columns: A sequence of 4 tactile images taken by the GelSight tactile sensor while the sensor was pressing (the first row), slipping over (the second row) and twisting (the third row) the fabric surface.
	}
	\label{fig:cloth}
\end{figure*}

We use the tactile data of the ViTac Cloth dataset we collected in~\cite{luo2018vitac}. The dataset includes 100 pieces of daily clothing of different materials, e.g., cotton, wool, ramie, silk and leather, and have different surface textures. Both data collected from a GelSight sensor and a digital camera is present in the dataset, however, only the tactile data is utilized as we focus on tactile texture recognition in this paper. The tactile data was collected by a GelSight sensor~\cite{yuan2017gelsight} while the cloth was lying flat. The sensor was held manually to interact with the surface of the fabrics. As the sensor moves on the cloth, a sequence of GelSight images of the cloth textures (raw video output from the camera) is recorded. On average each cloth was contacted by the sensor for around 30 times and the number of GelSight readings in each sequence ranges from 25 to 36. 
There were three different actions conducted to collect the tactile data for each piece of fabric, i.e., pressing, slipping and twisting. Three kinds of continuous sequences collected by the GelSight sensor under different operations (press, slip, twist) are shown in the Fig.~\ref{fig:cloth}, with their visual images illustrated only for visualisation purpose. 
In total, it contains 5,036 tactile images sequences with a ratio of $7:1:2$ for pressing, slipping and twisting.

\section{Experiments and analysis}\label{sec:experimentresults}
\subsection{Experimental Setup}
In our experiments, we aim to understand whether our designed spatio-temporal attention mechanisms can improve the performance of tactile texture recognition.
To this end, we conduct an ablation study with different neural network frameworks to learn how the spatial attention and temporal attention help the recognition task step by step.
We first apply a pre-trained AlexNet on the tactile sequences directly to form a baseline.
Subsequently, we add the spatial attention module to the AlexNet to learn the spatial features.
In a further step, both spatial attention and temporal attention modules are included to form the spatio-temporal attention model for tactile texture recognition. 

A short input length of the tactile sequence can reduce the processing time, but can also decrease the performance of recognition. To study the impact of the length of input sequences on our methods, we use different lengths for input sequences.
We only vary the length $n$ of a sequence ranging from $2$ to $7$, because the time complexity will be much higher whereas the improvement by attention mechanism is minimal if the length continues to increase over $7$.

We notice that there is data collected before the contact happens in the ViTac dataset and cannot provide any useful information for texture recognition. It is a quite common scene while robots interact with different objects to collect surface textures with a tactile sensor in a dynamic environment.
Therefore, we conduct two sets of experiments: 1) We detect the first valid tactile texture that indicates the sensor starting to contact the object surface during an operation in a tactile sequence, then we take the following $n$ consecutive tactile images after the detected texture as the input sequence and feed it to the network. 2) We include the tactile images before the contact as noisy data to verify the robustness of our model, i.e., 
we take $n$ consecutive tactile images starting from the very beginning in each tactile sequence of ViTac dataset, 
which is closer to an actual application.
This comparative experiment enables us to investigate how the noisy data affects the performance of our models.

We split the data by a $7:2:1$ ratio for training, validation and testing. 
We use Keras (Tensorflow backend) to implement our models. Due to GPU memory limitation, we include $10$ temporal attention modules running in parallel in the STAM.

\begin{table*}[htbp]
	\centering
		\caption{Texture recognition results using different models with different lengths of input sequences. }
		\label{tab:re1}
        \scalebox{1.25}{
		\begin{tabular}{c| c | c | c | c| c| c}
			\hline
			Models & $n=2$ & $n=3$ & $n=4$& $n=5$ & $n=6$& $n=7$ \\
			\hhline{=|=|=|=|=|=|=}
		CNNs & 67.23\% & 72.04\% & 75.26\% & 78.06\% & 79.56\% & 81.29\%\\
		CNNs+Spatial Attention & {72.12\%} & {73.97\%} & {78.60\%} & {80.43\%}  & {80.43\%}& {80.86\%}\\
		STAM & \pmb{76.50\%} & \pmb{79.35\%} & \pmb{80.00\%} & \pmb{80.64\%}  & \pmb{81.72\%}& \pmb{81.93\%}\\
		\hline
		\end{tabular}}
\end{table*}

\begin{table*}[htbp]
	\centering
		\caption{Texture recognition results while tactile images collected before the contact are introduced into the dataset.}
		\label{tab:re2}
        \scalebox{1.25}{
		\begin{tabular}{c| c | c | c | c| c| c}
			\hline
			Models & $n=2$ & $n=3$ & $n=4$& $n=5$ & $n=6$& $n=7$ \\
			\hhline{=|=|=|=|=|=|=}
		CNNs & 53.20\% & 58.20\% & 59.60\% & 61.23\% & 64.60\% & 69.40\%\\
		CNNs+Spatial Attention & {55.40\%} & {60.80\%} & {62.60\%} & {62.80\%}  & {65.40\%}& {71.00\%}\\
		STAM & \pmb{72.00\%} & \pmb{72.20\%} & \pmb{75.80\%} & \pmb{76.61\%}  & \pmb{80.80\%}& \pmb{80.20\%}\\
		\hline
		\end{tabular}}
\end{table*}

\subsection{Experimental Results}
From Table~\ref{tab:re1} and Table~\ref{tab:re2}, we can see that our proposed STAM model achieves the best recognition performance for all the cases with different lengths of sequences.
Compared to the baseline method which uses AlexNet, the recognition accuracy of our proposed STAM model has increased by $4.45\%$ on average. 
When the noisy data is introduced into the dataset, the performance of the STAM improves in a further step compared with the baseline, by $15.23\%$ on average. In addition, with the help of spatial attention and temporal attention, the recognition accuracy of the STAM model improves step by step in most of the cases with different lengths of sequences. Furthermore, we can see that the recognition accuracy has an upward tendency as the length of the sequence increases. 


From Table~\ref{tab:re1}, we can see that the recognition accuracies of all the models on texture recognition tasks improve as more tactile images exist in a sequence. 
The changes in the lengths of sequences have different effects on the performance of different methods.
Specifically, when the lengths of the sequences decrease from $7$ to $2$, the accuracy of CNNs based method falls 14.06\%, the accuracy of the spatial attention method falls 8.74\%, and the performance of proposed STAM only has 5.43\% drop.  
We can find that the CNNs based model obtains the worst performance for most of the cases. In terms of the model with spatial attention module, which is connected with fine-tuned AlexNet to unify spatial features, the accuracy has a slightly improvement compared with the CNNs only model. With both spatial attention and temporal attention applied, our proposed STAM model achieves the best performance for all of the cases. These results demonstrate that, by taking the advantage of the attention mechanism, it is able to efficiently extract salient information from each frame. 

Similar trend can be observed in Table~\ref{tab:re2}. All the models obtain their best performance with $7$ tactile images used in a sequence. From this table, we can find that, even though adding noisy data results in lower accuracy for all the models on the texture recognition task, the performance of our proposed STAM model still maintains at the same level while the other models cannot sustain the recognition accuracy compared with the accuracy in Table~\ref{tab:re1}. Specifically, in the column where \textbf{\textit{$n = 6$}}, the accuracy of the baseline model decreases by $14.96\%$ while the performance of the STAM model only decreases by $0.92\%$ after introducing the noisy data. 
In addition, when \textbf{\textit{$n = 2$}}, the STAM achieves the largest improvement, by $18.8\%$ compared with the baseline.
These results prove that our proposed STAM has strong robustness to a redundant dataset and can select informative information effectively from space and time dimensions.
We also calculate the number of parameters in the baseline model and STAM respectively. The baseline model contains $57.4$ million parameters and STAM contains $68.8$ million parameters ($n=2$). It demonstrates that our spatial-temporal attention mechanism is compact and effective.

In general, our proposed STAM model has two distinct advantages: 
1) A more robust method is achieved. Noise has less effect on the STAM compared with other methods, which means that the attention mechanism can help to select salient features and suppress the noise for the recognition.
2) The informative features can be learnt more effectively. The STAM achieves a larger improvement with a short input sequence compared with baseline approach, which shows the efficiency of our methods with limited data.

\begin{figure*}[t]
	\centering
	\includegraphics[width=1.9\columnwidth]{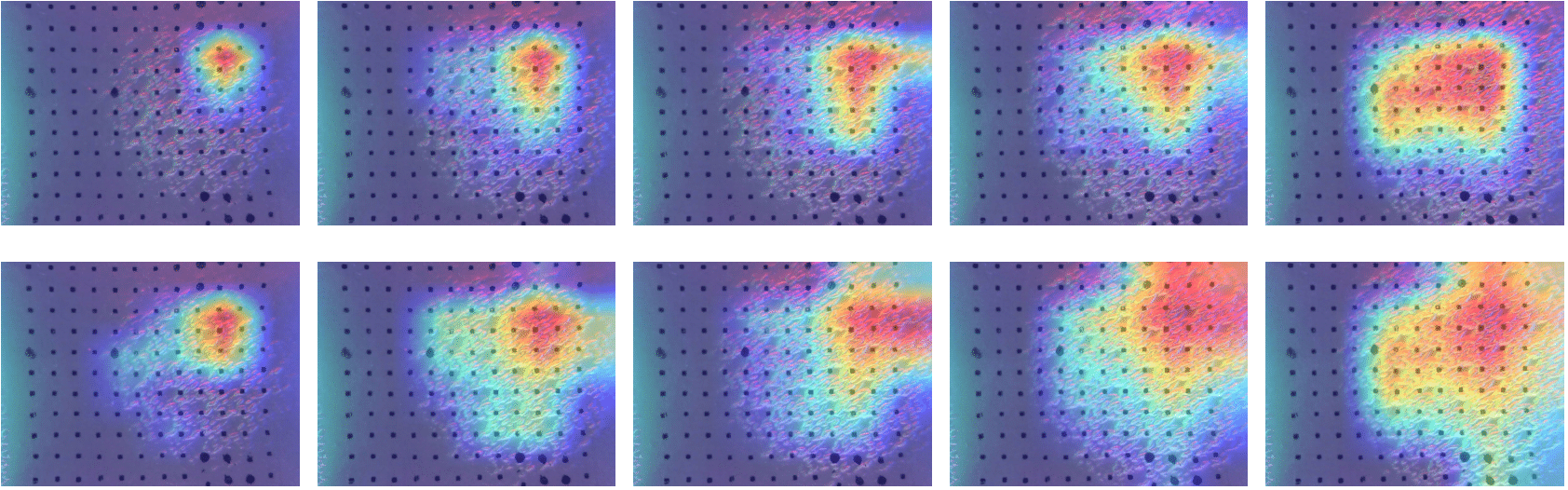}
	\caption{\textbf{\textit{Spatial Attention Distribution.}} 
    The background tactile images of the first row and second row are the same texture sequence collected while pressing against the fabric.
	The first and second rows represent the Grad-CAMs of non-attention and spatial attention models respectively. The lighter regions refer to the larger weights that are assigned, and vice versa. As seen from the results, more informative regions are activated by spatial attention mechanisms.} 
	\label{fig:heatmap}
\end{figure*}

\textit{\textbf{Spatial Attention Distribution.}} To ensure the effectiveness of spatial attention module, we visualize the gradient class activation maps (Grad-CAM)~\cite{selvaraju2017grad} on the non-attention method and spatial attention method respectively for the same texture sequence. The Grad-CAM uses the gradients to describe importance of location w.r.t. the classification result. It enables us to compare the differences of saliency regions activated by each method. As shown in Fig.~\ref{fig:heatmap}, our spatial attention mechanism makes more contact regions activated compared with the non-attention baseline, which means that the spatial attention module is able to extract more informative features for recognition.

\begin{figure*}[t]
	\centering
	\includegraphics[width=1.9\columnwidth]{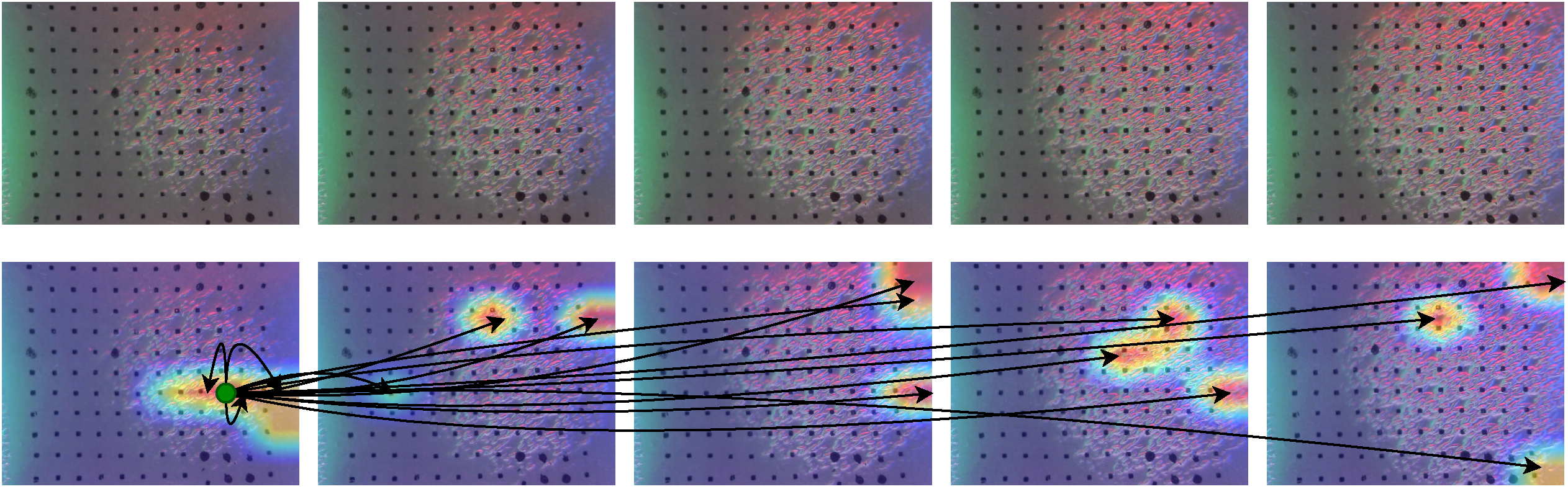}
	\caption{\textbf{\textit{Temporal Attention Distribution.}} Upper row: the raw tactile textures collected by pressing and sorted by time. Lower row: the temporal attention distribution where the green dot refers to the region to be updated, and the highlighted regions pointed by arrows in each tactile texture are $3$ most relevant regions that attend to synthesis of the green dot region. As seen in the highlighted regions, most of pointed parts by arrows are located in the latest added contact regions.     } 
	\label{fig:selfdistribution}
\end{figure*}

\textit{\textbf{Temporal Attention Distribution.}} To further investigate how the temporal attention performs in modeling long distance dependency, we visualize the attention map of our temporal attention module as shown in Fig.~\ref{fig:selfdistribution}. The value of attention map in temporal attention describes how much of a region attend to another region while updating its value. As we apply multiple temporal attention modules, the attention maps from each module are averaged to have an overall observation of correlations of any pairs of locations. Specifically, we choose a region in the first texture  which is marked as a green dot in Fig.~\ref{fig:selfdistribution}. Then we select 3 most relevant regions for each texture that are pointed by the black arrows. We find that the arrows always point to the latest added contact region in most cases which means that the green dot region is able to synthesize its value with the latest information through time in our proposed method.  




\section{CONCLUSIONS}
\label{sec:conclusion}
In this paper, we propose a novel Spatio-Temporal Attention Model (STAM) for tactile texture recognition,
which pays attention to both spatial focus of a tactile image and the temporal focus of a tactile sequence.
In the experiments on discriminating between 100 different fabric textures, the spatially and temporally selective attention has resulted in a significant improvement of the recognition accuracy, by up to 18.8\%, compared to the non-attention based models. After introducing the noisy data, our proposed STAM shows strong robustness over other methods. Furthermore, the STAM is able to select informative features efficiently with limited data through space and time dimensions. 
The improved tactile texture perception can be applied to facilitate robot tasks like grasping and manipulation. 

\section*{ACKNOWLEDGMENT}
This work was supported by the EPSRC project ``Robotics and Artificial Intelligence for Nuclear" (EP/R026084/1).



{\small
	\bibliographystyle{ieeetr}
	\bibliography{reference.bib}
}

\end{document}